# Title

Hardware implementation of timely reliable Bayesian decision-making using memristors

# Authors

Lekai Song[1], Pengyu Liu[1], Yang Liu[1,2], Jingfang Pei[1], Wenyu Cui[3], Songwei Liu[1], Yingyi Wen[1], Teng Ma[3], Kong-Pang Pun[1], Leonard W. T. Ng[4], Guohua Hu[1]

# Affiliations

[1]Department of Electronic Engineering, The Chinese University of Hong Kong, Shatin, New Territories, Hong Kong S. A. R., China
[2]Shun Hing Institute of Advanced Engineering, The Chinese University of Hong Kong, Shatin, New Territories, Hong Kong S. A. R., China
[3]Department of Applied Physics, Hong Kong Polytechnic University, Hung Hom, Kowloon, Hong Kong S. A. R., China
[4]School of Materials Science and Engineering, Nanyang Technological University, Singapore 639798, Singapore

*Correspondence to: Guohua Hu (ghhu@ee.cuhk.edu.hk)

# Abstract

Brains perform decision-making by Bayes theorem. The theorem quantifies events as probabilities and, based on probability rules, renders the decisions. Learning from this, Bayes theorem can be applied to enable efficient user-scene interactions. However, given the probabilistic nature, implementing Bayes theorem in hardware using conventional deterministic computing can incur excessive computational cost and decision latency. Though challenging, here we present a probabilistic computing approach based on memristors to implement the Bayes theorem. We integrate memristors with Boolean logics and, by exploiting the volatile stochastic switching of the memristors, realise probabilistic logic operations, key for hardware Bayes theorem implementation. To empirically validate the efficacy of the hardware Bayes theorem in user-scene interactions, we develop lightweight Bayesian inference and fusion hardware operators using the probabilistic logics and apply the operators in road scene parsing for self-driving, including route planning and obstacle detection. The results show our operators can achieve reliable decisions in less than 0.4 ms (or equivalently 2,500 fps), outperforming human decision-making and the existing driving assistance systems.

I.   Main

Brains are adept at making timely reliable decisions for events even with uncertainties. This ability is attributed to the *Bayes theorem*, a mathematical principle used to determine the probability of an event to occur when given certain conditions (*1*). In Bayes theorem, the events, conditions, and their relations are all quantified and computed as probabilities, following the fundamental probability rules, to render timely reliable decisions. As such, the Bayes theorem shows promise in visualizing the potential risks and decision confidence in addressing practical problems in, for instance, user-scene interactions (*2, 3*). However, given the probabilistic nature, implementing Bayes theorem can be challenging due to the lack of specialised hardware to perform probabilistic data representation and computation. Using the conventional deterministic computing systems for Bayes theorem implementation can inevitably incur excessive computational cost and decision latency (*4*).

Idealistic Bayes theorem implementation demands a probabilistic computing solution to realise the probabilistic data representation and computation. One feasible approach is to encode the data into streams of random bits, termed *stochastic numbers*, where the probability of the bit 1s determines the values (*5*). The stochastic numbers, when fed into standard Boolean logic gates, can undergo bitwise Boolean operations and yield reconstructed stochastic numbers as the computational results. Notably, the inputs and results obey statistical relations, thereby simplifying the logic operations (*5–7*) and facilitating lightweight Bayes theorem implementation. Early electronic circuits based on, e.g., *linear feedback shift registers* (LFSR), have demonstrated the ability in stochastic number encoding by exploiting the stochasticity inherent in the circuits (*8–10*). The circuits can also be conveniently integrated with Boolean logics to process the stochastic numbers conforming to probabilistic computing rules. However, intricate pre-/post-processing circuit designs are typically required to, for instance, ensure the results are corrupted by improper correlations between the stochastic numbers (*11, 12*). This can result in considerable computational costs.

Herein, to address the problem, we present a memristor-enabled probabilistic computing approach for hardware implementation of Bayes theorem. We integrate volatile memristors with Boolean logics and, by exploiting the volatile stochastic switching of the memristors, realise probabilistic data encoding and logic operations that are essential for Bayes theorem implementation. Using the memristor-enabled probabilistic logics, we design lightweight Bayesian hardware operators that are capable of performing two key Bayesian functions, that are inference and fusion. The hardware operators are applied in road scene parsing for self-driving, including route planning and obstacle detection, to empirically validate their effectiveness in practical user-scene interactions. Our results show the operators can perform reliable decisions in less than 0.4 ms (or equivalently 2,500 fps). This outperforms human decision-making and the advanced driver assistance systems, underpinning the potential to enable efficient Bayesian user-scene interactions in, for instance, self-driving, virtual reality, and robotics.

II.   Results

**Volatile memristors**

Memristors are resistive switching devices, and they can present varying volatility depending on the switching mechanisms (*13*). Amongst them, the volatile memristors behave with self-reset threshold switching (*14*). Besides, memristors tend to exhibit nonideal stochasticity in the switching, due to their underlying stochastic switching dynamics. For example, as a result of the stochastic diffusion of the conductive elements, filamentary memristors switch with stochasticity (*15*). The stochasticity, however, along with the self-reset threshold switching behaviour, can potentially enable the volatile memristors to probabilistically encode data with reduced hardware and computational cost (*16–18*).

Towards probabilistic data representation for hardware Bayes theorem implementation, we develop volatile filamentary memristors from solution-processed hexagonal boron nitride (hBN), following our previous report (*19*). We start with liquid-phase exfoliation of hBN, and then fabricate memristors with active silver metal electrode via photolithography (Fig. S1). Figure 1a shows an array of the fabricated memristors. To assess the switching behaviour, we conduct sweeping test of the memristors (Fig. 1b). In a typical switching, the memristor switches to a low resistive state when the bias is over a threshold voltage $V_{th}$, and spontaneously resets to a high resistive state when the bias recedes below a hold voltage $V_{hold}$. The switching is governed by the underlying silver ion diffusion and filament formation/rupture, and the exhibited volatility arises from the interplay between the bias and the Joule heat – a bias below $V_{hold}$ cannot sustain the silver filaments due to the accompanying Joule heat (*19*). See also Fig. S2 for the transient switching of the memristors, with a switching time of ~50 ns, a relaxation time of ~1,100 ns, and a switching energy of ~0.16 nJ.

As the sweeping test continues, the memristor exhibits consistent cycle-to-cycle stochasticity in $V_{th}$ (2.08 ± 0.28 V) and $V_{hold}$ (0.98 ± 0.30 V) in the switching, well-fitting Gaussian distributions. The stochasticity arises from the underlying switching dynamics of the memristors, as the silver ion diffusion and filament formation/rupture are stochastic (*15*, *20*, *21*). To assess the stochasticity further, we perform a sampling test with 10 randomly selected devices from the array (Fig. S3). As demonstrated, all the 10 sampled devices exhibit a stable stochasticity in the switching. We plot the overall device-to-device stochasticity in Fig. 1c, with the distributions of the measured $V_{hold}$ and $V_{th}$ of the individual sampled memristors in Fig. 1d. The device-to-device stochasticity is ~8% in $V_{th}$, measured by the *coefficient of variation*, i.e., the ratio of the standard deviation to the mean of $V_{th}$ in sampled memristors, proving a high device-to-device uniformity. Having confirmed the stable, uniform cycle-to-cycle and device-to-device stochastic switching, we perform stochastic process modelling on $V_{th}$. The measured $V_{th}$ of each of the sampled memristors follows a mean-reverting behaviour with random fluctuations, conforming to the Ornstein-Uhlenbeck process (Fig. S4). Ornstein-Uhlenbeck process describes a stochastic process in a dynamical system (*16*). This indicates the stability of the switching stochasticity of our memristors in long-term switching. Indeed, the pulsed cycling switching test proves the switching endurance of our memristors over $10^6$ cycles (Fig. 1e).

The demonstrated stable volatile stochastic switching suggests that our memristors when exploited in probabilistic computing can encode the inputs into desired stochastic numbers in lightweight, where the volatile switching can eliminate resetting circuits and operations, while the switching stochasticity can be harnessed to facilitate stochastic number encoding. On the other hand, importantly, the device-to-device uniformity and high fabrication yield as proved in sampling test (Fig. S3) can facilitate simplified circuit design and calibration in practical probabilistic computing hardware implementation.

## Probabilistic logics

Using our volatile memristors, we then design stochastic number encoders (SNEs; Fig. 2a) for encoding stochastic numbers with desired probabilities and correlations, to realise probabilistic data representation as required by Bayes theorem. Here the memristors are integrated with comparators to implement lightweight SNEs. See also Fig. S5 for the hardware implementation. When fed with pulsed signal $V_{in}$, the memristors are switched stochastically, and the output from the memristors carrying the stochasticity is binarized by the comparators via the reference $V_{ref}$. This then yields stochastic numbers with probabilities, and the probabilities can be well-regulated by $V_{in}$ and $V_{ref}$. Notably, as designed, the stochastic numbers from an SNE are correlated, while those from two or more parallel SNEs are uncorrelated. Figure 2b presents the probability of uncorrelated stochastic numbers $P_{uncorrelated}$ with respect to $V_{in}$. As $V_{in}$ increases, $P_{uncorrelated}$ is increased, as the memristors tend to switch on with a higher probability. This relation follows a sigmoidal fit, proving that $V_{in}$ can effectively regulate $P_{uncorrelated}$. Similarly, Fig. 2c presents the probability of correlated stochastic numbers $P_{correlated}$ with respect to $V_{ref}$. In this case, $P_{correlated}$ is decreased as $V_{ref}$ increases, as $V_{ref}$ serves as the threshold for binarization. This relation also follows a sigmoidal fit, proving that $V_{ref}$ can also effectively regulate $P_{correlated}$. Therefore, as demonstrated, our SNEs can perform stochastic number encoding with desired and well-regulated probabilities and correlations, suggesting that our SNEs can facilitate integration with Boolean logics for performing probabilistic computing.

As examples, here we show the integration of our SNEs with standard Boolean logics to build probabilistic AND and MUX in uncorrelation, as they can conduct key Boolean operations for Bayes theorem (Fig. 2d). For probabilistic AND, two parallel SNEs are connected to an AND gate, and the uncorrelated stochastic numbers from the SNEs serve as the inputs to the AND gate. Upon operation, based on the $P_{uncorrelated}$-$V_{in}$ relation in Fig. 2b, the SNEs when fed with pulsed signals output uncorrelated stochastic numbers, denoted as $a$ and $b$, with probabilities of $P(a)$ and $P(b)$, respectively. $a$ and $b$ are then bit-by-bit fed into the AND gate to output a stochastic number, denoted as $c$, with a probability of $P(c)$. We show in Fig. 2e the corresponding stochastic numbers and probabilities from the hardware test. The statistical relation, i.e., $P(a)P(b) \approx P(c)$, proves that the probabilistic AND successfully functions as a multiplier to conduct one-step multiplication of the stochastic numbers. Here we note that, compared to the conventional binary multipliers, the probabilistic AND not only simplifies the circuit design but also reduces the computational cost. The probabilistic AND can also be configured in correlation. In this case, SNEs are configured to output correlated stochastic numbers based on the $P_{correlated}$-$V_{in}$ relation in Fig. 2c, and the correlated probabilistic AND outputs the minimum of the inputs i.e., $P(c) \approx min(P(a), P(b))$ (Fig. 2e). We also design probabilistic MUX in both uncorrelation and correlation, and it performs one-step weighted addition (Fig. 2e). Here the select and inputs of the probabilistic MUX must be uncorrelated to perform the operations properly (Fig. S6). We present in Table S1 a summary of the probabilistic logics including AND, OR, XOR, and MUX in varying correlations.

The probabilistic logics with the desired probabilities and correlations can potentially be integrated to enable probabilistic computing towards Bayes theorem implementation. Note that in the above demonstrations, the stochastic numbers are all encoded in 100-bit length for illustrative purposes. A longer bit length renders a higher precision in data representation, however, with a higher computational cost. The bit length can be adjusted to accommodate tasks with different precision requirements, given the trade-off between the computational cost and precision.

**Hardware implementation of Bayesian inference**

To instantiate Bayes theorem, we apply our probabilistic logics to implement Bayesian inference, a key decision-making method. Inference draws conclusions based on known facts. Despite investigated in multiple disciplines including logic, neuroscience, and computer science, the essence of inference still remains elusive (*22–24*). Nevertheless, the Bayes theorem provides a probabilistic solution to achieve inference (*25*). In Bayesian inference, *prior knowledge* denotes the known facts, and *belief* the confidence (or reliability) of a certain decision. Importantly, as the prior knowledge and beliefs are quantified as probabilities, Bayesian inference can effectively revise the prior knowledge with the new information, based on the rules of probability, to make reliable decisions.

Our lightweight memristor-based probabilistic logics are proven capable of performing probabilistic data representation and computation, conforming to Bayesian inference. Herein, we design a hardware operator using our probabilistic logics to perform Bayesian inference (Fig. 3a; see also Fig. S7 for the hardware implementation), following the Bayesian inference theorem:

$$P(A|B) = \frac{P(A)P(B|A)}{P(B)} = \frac{P(A)P(B|A)}{P(A)P(B|A) + P(\neg A)P(B|\neg A)}. \tag{1}$$

As designed, the probabilistic AND and MUX are integrated to conduct the one-step multiplication and weighted addition operations, respectively, for functioning as the numerator and denominator in equation (1). The division is achieved with a probabilistic MUX plus a D-Flip-Flop, following a classic divider design, CORDIV (*26*). When performing Bayesian inference, the prior knowledge with a prior probability $P(A)$ is updated when the new information with a marginal probability $P(B)$ is available, to yield a reliable decision with a posterior probability $P(A|B)$.

To investigate the effectiveness of our Bayesian inference operator in practical applications, we apply the operator in route planning for self-driving (Fig. 3a). Route planning has been a challenging topic in self-driving, demanding great efforts to deal with complex traffic, vehicle driving, and lane-changing conditions. Particularly, given that any delay in decision making can lead to severe traffic accidents, achieving timely reliable decisions is critical in route planning. Bayesian inference has been considered promising for performing route planning (*27*), as it can potentially make reliable decisions with prior knowledge (e.g. the traffic rules, road structures, and driving behaviours) and the latest lane information (e.g. the obstacles, accidents, and weather conditions). Here we assume a vehicle (red) considering changing its lane, with probability $P(A)$ representing its initial belief to cut in the lane, based on the prior knowledge, i.e. the traffic conditions before lane changing. To ensure a safe lane change, the vehicle may evaluate the situation in the target lane to determine whether to cut in through Bayesian inference. In this case, $P(B)$ represents a marginal probability related to the latest lane condition, i.e. an incoming vehicle (blue) on the target lane. With $P(A)$ and $P(B)$, the vehicle may update its belief based on the Bayesian inference theorem and get a posterior probability $P(A|B)$. For illustration, $P(A) = 57\%$ and $P(B) = 72\%$ are assumed here for Bayesian inference (Fig. 3b). The operator integrates the prior knowledge with the latest lane information to make a final decision with a probability $P(A|B) = 63\%$. This probability aligns with the theoretical result in equation (1), ~61%, validating the effectiveness of our hardware operator in performing Bayesian inference. With the updated belief, i.e. $P(A) < P(A|B)$, the vehicle may make a lane-changing decision with a higher reliability. However, here we must note the decision-making is essentially not limited to lane-changing. For example, when $P(A) > P(A|B)$, the hardware operator can make a decision for maintaining its current lane with a higher reliability.

Note that, to make our hardware operator lightweight for timely decision-making, we maximise the sharing of the SNEs. This is possible with a well-regulation of the probabilities and correlations of the stochastic numbers that ensures the proper functions of the Multiplier, Adder, and Divider. To clarify, we quantify pairwise correlations between the stochastic numbers involved in the Bayesian inference using *Pearson correlation* and *SC correlation* (Fig. 3c and d). The results confirm that all the probabilistic logics in our operator work in the desired correlations and, as such, conduct the desired Boolean operations for performing Bayesian inference. Though our route planning studies the case where the decision-making is governed by one variable only (denoted as one-parent-one-child, i.e. A→B), our operator can be readily generalised to more cases, for instance, two-parent-one-child ($A_1$→B←$A_2$) and one-parent-two-child ($B_1$←A→$B_2$) (Fig. S8). In addition, arising from the fast switching of the memristors (<4 μs in total per bit, Fig. S2), our Bayesian inference operator with 100-bit stochastic number encoding can achieve Bayesian inference with less than 0.4 ms per frame, or equivalently 2,500 fps. Note that here we neglect the delays of the comparators and Boolean logics, as the switching of the memristors is the bottleneck of the operator. In this consideration, our operator outperforms human decision-making (0.7-1.5 ms reaction time (*28*)) and the advanced driver assistance systems (30-45 fps (*29*)) in route planning.

**Hardware implementation of Bayesian fusion**

Besides inference, fusion is another key Bayesian decision-making. While inference integrates the past and present information, fusion considers the information from multiple modalities. Given that each modality represents fragmented information only, Bayesian fusion fuses multimodal information to address the single-modal shortages and as such, achieve more reliable decisions (*22*). As with inference, the information in Bayesian fusion is represented and computed by probabilities and can lead to excessive computational cost and latency. Here we investigate the potential of our probabilistic logics in implementing Bayesian multimodal fusion in hardware. We design an operator following the Bayesian fusion theorem (Fig. 4a). See also S9 for the hardware implementation.

To investigate the effectiveness of our hardware Bayesian fusion operator in practical applications, we apply the operator in obstacle detection in self-driving (Fig. 4a). Here we assume a self-driving vehicle equipped with RGB and thermal cameras, with the RGB camera to extract the obstacle signatures mainly in good visibility and the thermal camera obstacle signatures only with heat emission. We test with a real dataset, *FLIR* (*30*), that consists of aligned RGB-thermal images captured at different visibility conditions. Pre-trained edge networks tailored for processing RGB and thermal signals are used to output single-modal detection decisions, denoted as $P(y|x_1)$ and $P(y|x_2)$, where $y$ represents the detected obstacles, and $x_1$ and $x_2$ the RGB and thermal input. Considering the limitations of the RGB and thermal cameras, the single-modal decisions tend to be unreliable. As shown (Fig. 4b), the thermal camera loses certain obstacles, as a result of insufficient thermal emissions from the obstacles or received by the camera. RGB camera also misses obstacles, particularly during low-visibility nighttime. Besides, these two single modalities sometimes fail to provide confident decisions.

Given this, RGB-thermal Bayesian fusion has been considered promising for realizing reliable decision-making, particularly, during low-visibility nighttime, rain, and fog (*2*, *3*). Our Bayesian fusion operator can fuse the single-modal RGB and thermal decisions into $p(y|x_1, x_2)$, following,

$$p(y|x_1, x_2) = \frac{p(x_1, x_2|y)p(y)}{p(x_1, x_2)} \propto p(x_1, x_2|y)p(y) \tag{2}$$

Assuming $x_1$ and $x_2$ are conditionally independent with given y, we have,

$$p(x_1, x_2|y) = p(x_1|y)p(x_2|y) \tag{3}$$

Substituting equation (3) into (2), we get the multimodal fusion decision,

$$p(y|x_1, x_2) \propto p(x_1|y)p(x_2|y)p(y) = \frac{p(x_1|y)p(y)p(x_2|y)p(y)}{p(y)} \propto \frac{p(y|x_1)p(y|x_2)}{p(y)} \tag{4}$$

The RGB-thermal Bayesian fusion results prove our operator is effective in addressing the single-modal detection issues (Fig. 4b). On the one hand, our operator can integrate the single-modal RGB and thermal decisions to generate a final decision, resolving the issue of target missing. On the other hand, our operator can make more confident decisions by leveraging the Bayesian fusion theorem.

As with the Bayesian inference operator, our Bayesian fusion operator also maximises the sharing of the SNEs and makes good use of the correlations to realise lightweight circuit design. Besides, equation (4) can be easily generalised, according to Ref. (*31*), to allow more modal fusion by,

$$p(y|x_1, x_2 \cdots x_M) = \frac{\prod_{i=1}^{M} p(y|x_i)}{p(y)^{M-1}} \tag{5}$$

Similarly, arising from the fast switching of the memristors, our Bayesian fusion operator with 100-bit stochastic number encoding can also easily achieve fusion with less than 0.4 ms per frame (equivalently 2,500 fps). Considering the sampling rate of the cameras (10-30 fps (*32*)) and the processing speed of the pre-trained networks (300 fps (*33*)) deployed in real-world self-driving, our operator is a promising timely reliable obstacle detection solution for self-driving.

## III. Discussion

We have implemented a memristor-enabled probabilistic computing approach for performing timely reliable Bayesian decision-making in hardware. We develop lightweight probabilistic logics using volatile stochastic memristors and achieve probabilistic data representation and logic operations to facilitate hardware Bayes theorem implementation. As practical applications of the hardware Bayes theorem, we design Bayesian inference and fusion operators using the probabilistic logics and prove their remarkable performance in road scene parsing for self-driving, including route planning and obstacle detection.

Given the demonstrated remarkable Bayesian decision-making performance and the compacity of the circuit designs, as well as the scalability of our memristors, our hardware Bayes theorem approach is readily scaled up and generalised to develop Bayesian systems for real-world user-scene interactions. To explore the feasibility, we show via simulation that a large-scale Bayesian fusion can process high-throughput road scene videos (Movie S1). As demonstrated, the fusion resolves the target missing issue, with significantly higher chances to detect the obstacles than the single-modal thermal (by 85%) or RGB (by 19%). For example, the running child obscured by the harsh lighting is hardly detected by RGB. Besides, importantly, the fusion achieves decisions at a higher confidence. Arising from the fast switching of the memristors, the large-scale Bayesian fusion can achieve user-scene interactions with a response time of less than 0.4 ms, or equivalently over 2,500 fps, fulfilling the requirements of widespread applications including self-driving, virtual reality, robotics, and beyond. Though promising, realising large-scale Bayesian systems requires high-level integration of the memristors with logic circuits, and parallel operations of the large-scale circuits. Hardware and algorithm codesigns are also needed to address or accommodate the non-idealities, e.g. noises and delays from the circuits.

## IV. Methods

**Memristors:** Pristine hBN powder and all chemicals are purchased from Sigma-Aldrich and used as received. Liquid-phase exfoliation, ink formulation, and deposition of hBN follow our method previously reported in Ref. (*19*). In a typical process (Fig. S1), the memristor is fabricated in a vertical Pt/Au/hBN/HfO$_x$/Ag configuration, where hBN is deposited by slot-die coating, the HfO$_x$ layer (20 nm) is deposited by atomic layer deposition, and the metal electrodes (5/15 nm Pt/Au and 30 nm Ag) are patterned by photolithography and deposited by electron beam evaporation. During device fabrication, the hBN layer after deposition is baked at 200°C for 2 hours.

**Stochastic number encoders and probabilistic logics:** To build the SNEs and probabilistic logics, the memristors are tested on a probe station and connected to the logic gates and other electronic devices on a breadboard. Tektronix Keithley 4200A-SCS parameter analyser with pulse measure units is used to measure the electrical characteristics of the memristors. Siglent arbitrary waveform generators and digital storage oscilloscopes are used to output signals and measure waveforms. To endow the stochastic numbers with a certain probability, based on the $P_{\text{uncorrelated}}$-$V_{\text{in}}$ and $P_{\text{correlated}}$-$V_{\text{in}}$ relations in Fig. 2b and c, each stochastic number encoder is fed with $n$-cycle pulsed signals of the corresponding $V_{\text{in}}$ to encode $n$-bit stochastic numbers.

**Stochastic number correlation:** The correlation between the stochastic numbers $S_x$ and $S_y$ is quantified using the *Pearson correlation* ($\rho$) and *SC correlation* ($SCC$) (*11*), following,

$$\rho(S_x, S_y) = \frac{ad - bc}{\sqrt{(a+b)(a+c)(b+d)(c+d)}}$$

and

$$SCC(S_x, S_y) = \begin{cases} \dfrac{ad - bc}{(a+b+c+d)\min(a+b, a+c) - (a+b)(a+c)}, & ad \geq bc \\ \dfrac{ad - bc}{(a+b)(a+c) - (a+b+c+d)\max(a-d, 0)}, & ad < bc \end{cases}$$

where $a$, $b$, $c$, and $d$ represent the counts of 1-1, 1-0, 0-1, and 0-0 pairs of two stochastic numbers $S_x$ and $S_y$, respectively.

**Bayesian fusion:** The pre-trained RGB-only and thermal-only networks used are the *Ultralytics YOLOv8* (*34*) and *Roboflow flir-data-set/22* (*30*), respectively. The prior of the obstacle $p(y)$ is assumed to be uniform for the convenience of circuit designs. Limited by the scalability of the lab-based realisation of the hardware Bayesian operator, the hardware Bayesian fusion is conducted on individual RGB-thermal image pairs only. The large-scale Bayesian fusion on videos as presented in Movie S1 is conducted via simulation in Python. Considering the output probability may exceed one, a normalization module is integrated to ensure reasonable outputs as the final multimodal fusion decisions (Fig. S10).

## References

1. D. C. Knill, A. Pouget, The Bayesian brain: The role of uncertainty in neural coding and computation. *Trends in Neurosciences* **27**, 712–719 (2004).
2. W. Zhou, X. Lin, J. Lei, L. Yu, J. N. Hwang, MFFENet: multiscale feature fusion and enhancement network for RGB-thermal urban road scene parsing. *IEEE Transactions on Multimedia* **24**, 2526–2538 (2022).
3. Y. Sun, W. Zuo, P. Yun, H. Wang, M. Liu, FuseSeg: semantic segmentation of urban scenes based on RGB and thermal data fusion. *IEEE Transactions on Automation Science and Engineering* **18**, 1000–1011 (2021).
4. D. Bonnet, T. Hirtzlin, A. Majumdar, T. Dalgaty, E. Esmanhotto, V. Meli, N. Castellani, S. Martin, J.-F. Nodin, G. Bourgeois, J.-M. Portal, D. Querlioz, E. Vianello, Bringing uncertainty quantification to the extreme-edge with memristor-based Bayesian neural networks. *Nature Communications* **14**, 7530 (2023).
5. A. Alaghi, W. Qian, J. P. Hayes, The promise and challenge of stochastic computing. *IEEE Transactions on Computer-Aided Design of Integrated Circuits and Systems* **37**, 1515–1531 (2018).
6. R. K. Budhwani, R. Ragavan, O. Sentieys, Taking advantage of correlation in stochastic computing. *Proceedings - IEEE International Symposium on Circuits and Systems*, 7–10 (2017).
7. A. Alaghi, J. P. Hayes, Survey of Stochastic Computing. *ACM Transactions on Embedded Computing Systems* **12**, 1–19 (2013).
8. P. Knag, W. Lu, Z. Zhang, A native stochastic computing architecture enabled by memristors. *IEEE Transactions on Nanotechnology* **13**, 283–293 (2014).
9. S. Gaba, P. Sheridan, J. Zhou, S. Choi, W. Lu, Stochastic memristive devices for computing and neuromorphic applications. *Nanoscale* **5**, 5872–5878 (2013).
10. S. A. Salehi, Low-Cost Stochastic Number Generators for Stochastic Computing. *IEEE Transactions on Very Large Scale Integration (VLSI) Systems* **28**, 992–1001 (2020).
11. A. Alaghi, J. P. Hayes, Exploiting correlation in stochastic circuit design. *2013 IEEE 31st International Conference on Computer Design, ICCD 2013*, 39–46 (2013).
12. H. Ichihara, S. Ishii, D. Sunamori, T. Iwagaki, T. Inoue, Compact and accurate stochastic circuits with shared random number sources. *2014 32nd IEEE International Conference on Computer Design, ICCD 2014*, 361–366 (2014).
13. L. O. Chua, Memristors on 'edge of chaos.' *Nature Reviews Electrical Engineering* **1** (2024).
14. S. Kumar, X. Wang, J. P. Strachan, Y. Yang, W. D. Lu, Dynamical memristors for higher-complexity neuromorphic computing. *Nature Reviews Materials* **7**, 575–591 (2022).
15. J. Tang, F. Yuan, X. Shen, Z. Wang, M. Rao, Y. He, Y. Sun, X. Li, W. Zhang, Y. Li, B. Gao, H. Qian, G. Bi, S. Song, J. J. Yang, H. Wu, Bridging Biological and Artificial Neural Networks with Emerging Neuromorphic Devices: Fundamentals, Progress, and Challenges. *Advanced Materials* **31**, e1902761 (2019).
16. S. Dutta, G. Detorakis, A. Khanna, B. Grisafe, E. Neftci, S. Datta, Neural sampling machine with stochastic synapse allows brain-like learning and inference. *Nature Communications* **13**, 2571 (2022).
17. W. A. Borders, A. Z. Pervaiz, S. Fukami, K. Y. Camsari, H. Ohno, S. Datta, Integer factorization using stochastic magnetic tunnel junctions. *Nature* **573**, 390–393 (2019).
18. Q. Xia, J. J. Yang, Memristive crossbar arrays for brain-inspired computing. *Nature Materials* **18**, 309–323 (2019).
19. L. Song, P. Liu, J. Pei, F. Bai, Y. Liu, S. Liu, Y. Wen, L. W. T. Ng, K. P. Pun, S. Gao, M. Q. H. Meng, T. Hasan, G. Hu, Spiking neurons with neural dynamics implemented using stochastic memristors. *Advanced Electronic Materials* **2300564**, 1–9 (2023).
20. Z. Wang, S. Joshi, S. E. Savel'ev, H. Jiang, R. Midya, P. Lin, M. Hu, N. Ge, J. P. Strachan, Z. Li, Q. Wu, M. Barnell, G. L. Li, H. L. Xin, R. S. Williams, Q. Xia, J. J. Yang, Memristors with diffusive dynamics as synaptic emulators for neuromorphic computing. *Nature Materials* **16**, 101–108 (2017).
21. C. P. Hsiung, H. W. Liao, J. Y. Gan, T. B. Wu, J. C. Hwang, F. Chen, M. J. Tsai, Formation and instability of silver nanofilament in Ag-based programmable metallization cells. *ACS Nano* **4**, 5414–5420 (2010).
22. U. Noppeney, Perceptual Inference, Learning, and Attention in a Multisensory World. *Annual Review of Neuroscience* **44**, 449–473 (2021).
23. T. Rohe, A. C. Ehlis, U. Noppeney, The neural dynamics of hierarchical Bayesian causal inference in multisensory perception. *Nature Communications* **10**, 1–17 (2019).
24. Y. Lecun, Y. Bengio, G. Hinton, Deep learning. *Nature* **521**, 436–444 (2015).
25. R. van de Schoot, S. Depaoli, R. King, B. Kramer, K. Märtens, M. G. Tadesse, M. Vannucci, A. Gelman, D. Veen, J. Willemsen, C. Yau, Bayesian statistics and modelling. *Nature Reviews Methods Primers* **1** (2021).
26. T. H. Chen, J. P. Hayes, Design of division circuits for stochastic computing. *Proceedings of IEEE Computer Society Annual Symposium on VLSI, ISVLSI* **2016-Septe**, 116–121 (2016).
27. K. Wang, Y. Wang, B. Liu, J. Chen, Quantification of Uncertainty and Its Applications to Complex Domain for Autonomous Vehicles Perception System. *IEEE Transactions on Instrumentation and Measurement* **72**, 1–17 (2023).
28. M. Green, "How Long Does It Take to Stop?" Methodological Analysis of Driver Perception-Brake Times. *Transportation Human Factors* **2**, 195–216 (2000).
29. D. Gehrig, D. Scaramuzza, Low-latency automotive vision with event cameras. *Nature* **629**, 1034–1040 (2024).
30. T. Imaging, FLIR data set Dataset, *Roboflow Universe* (2024). https://universe.roboflow.com/thermal-imaging-0hwfw/flir-data-set.
31. Y.-T. Chen, J. Shi, Z. Ye, C. Mertz, D. Ramanan, S. Kong, "Multimodal Object Detection via Probabilistic Ensembling" in *Computer Vision -- ECCV 2022*, S. Avidan, G. Brostow, M. Cissé, G. M. Farinella, T. Hassner, Eds. (Springer Nature Switzerland, Cham, 2022), pp. 139–158.
32. D. J. Yeong, G. Velasco-hernandez, J. Barry, J. Walsh, Sensor and sensor fusion technology in autonomous vehicles: A review. *Sensors* **21**, 1–37 (2021).

33. H. Wang, C. Liu, Y. Cai, L. Chen, Y. Li, YOLOv8-QSD: An Improved Small Object Detection Algorithm for Autonomous Vehicles Based on YOLOv8. *IEEE Transactions on Instrumentation and Measurement* **73**, 1–16 (2024).
34. G. Jocher, J. Qiu, A. Chaurasia, Ultralytics YOLO, version 8.0.0 (2023); https://github.com/ultralytics/ultralytics.

# Figures

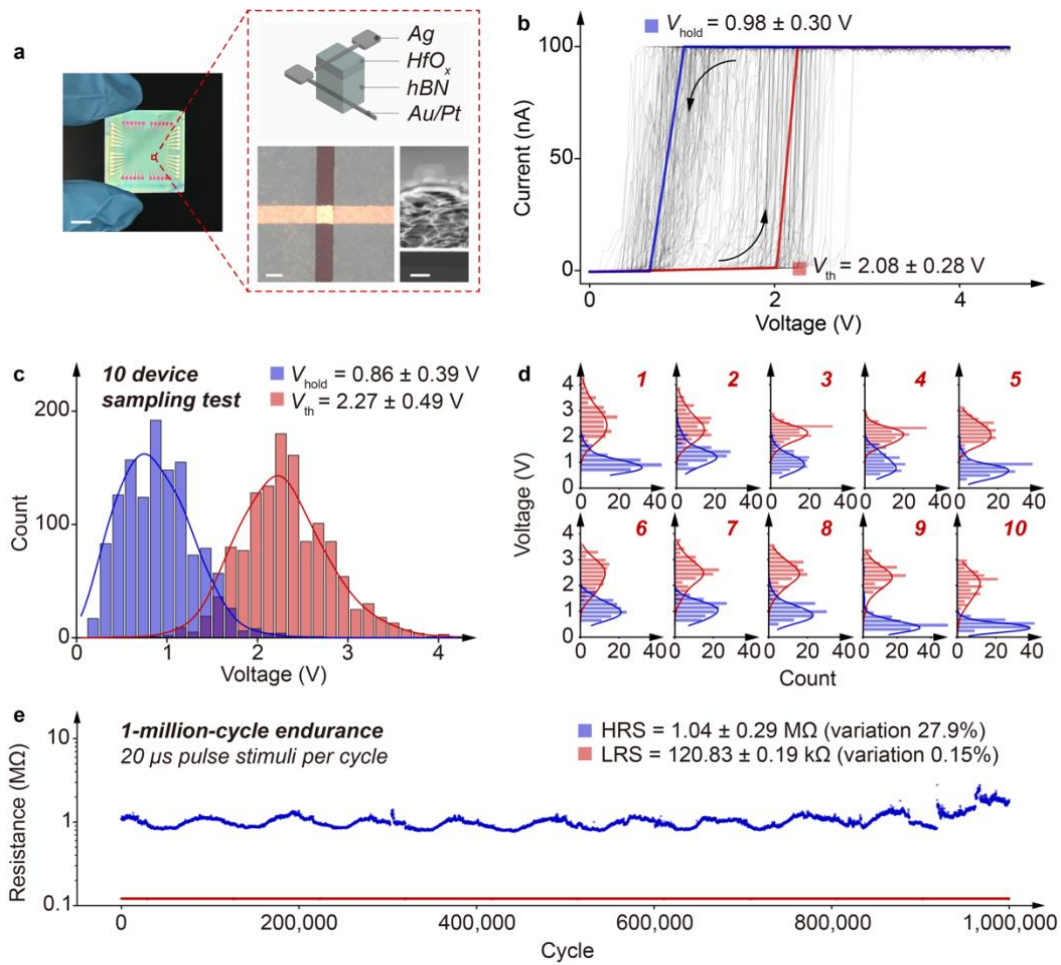

**Figure 1. Volatile switching memristors.** (a) 12 × 12 hBN memristors in a crossbar configuration, and the optical microscopic and cross-sectional transmission electron microscopic images of a typical memristor. Device area ~20 × 20 μm$^2$. Scale bars – 5 mm, 20 μm, and 100 nm. (b) Current-voltage output from a typical memristor, showing 128-cycle stochastic yet stable switching with a switching ratio of ~10$^5$. A compliance current of 100 nA is set. (c) Overall and (d) individual distributions of the measured $V_{hold}$ and $V_{th}$ of 10 randomly selected memristors in (a), along with the corresponding Gaussian fittings. 128 cycles are tested for each memristor. (e) Endurance test of a typical memristor undergoing 10$^6$ consecutive test cycles under pulsed stimuli. For each test cycle, a 20 μs voltage pulse of 10 V is set to fully program the memristor and an 80 μs voltage pulse of 0.1 V is set to read the output. The output of the memristor is amplified by an operational amplifier and measured by an oscilloscope. The high and low resistance states in each test cycle are computed based on the oscilloscope measurement, and plotted in blue and red, respectively. Both the states remain stable throughout the endurance test.

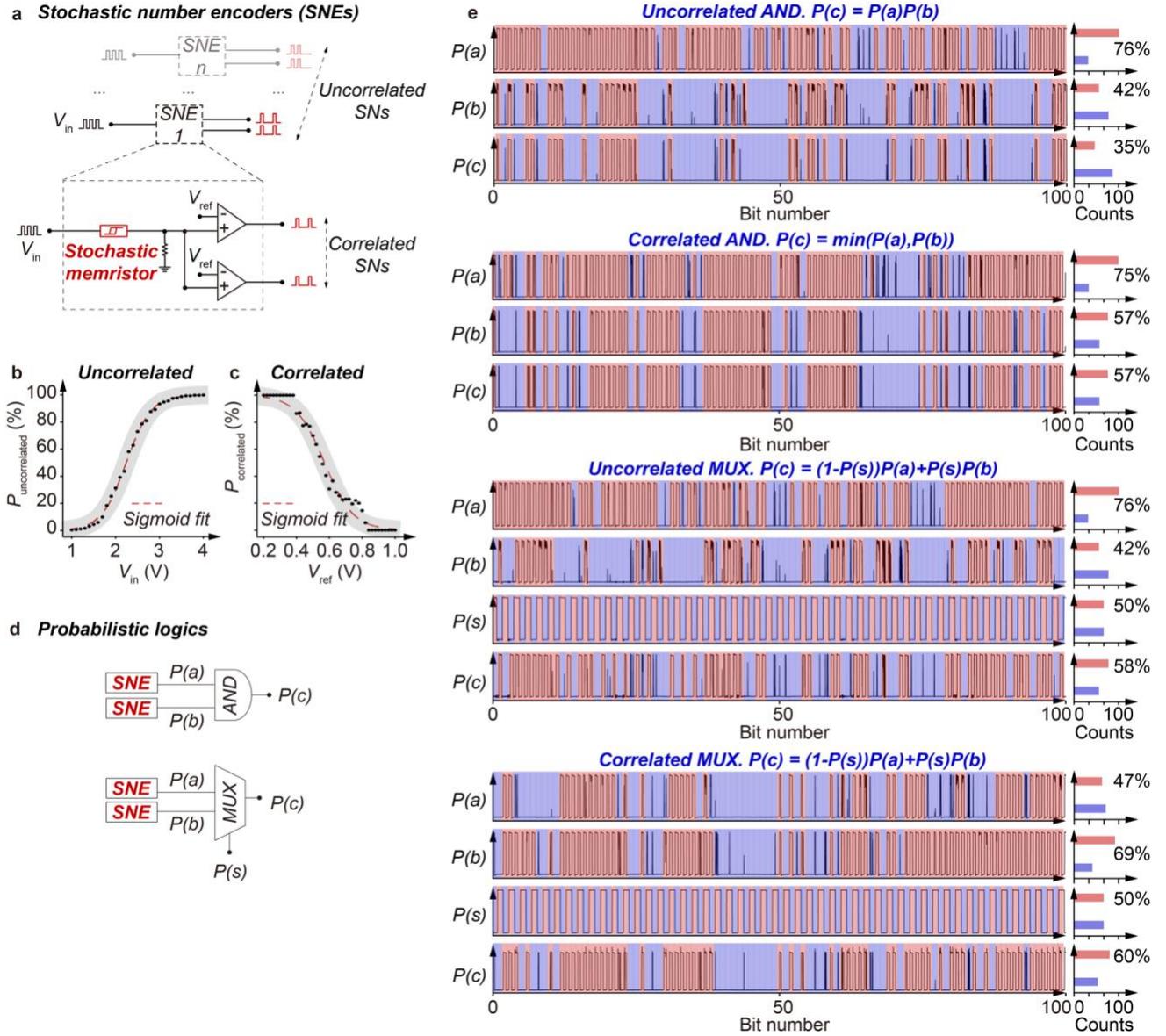

**Figure 2. Memristor-enabled probabilistic logics.** (a) Stochastic number encoders (SNEs), consisting of a memristor and a set of comparators. The output probability and correlation are regulated by the input $V_{in}$ and reference $V_{ref}$. An SNE via circuit designs can output correlated stochastic numbers, while two or more parallel SNEs can output uncorrelated stochastic numbers. (b) $P_{uncorrelated}$-$V_{in}$ and (c) $P_{correlated}$-$V_{ref}$ relations of the SNEs in uncorrelation and correlation, well-fitting sigmoid functions $P_{uncorrelated} = 1/(1 + exp[-3.56(V_{in} - 2.24)])$ and $P_{correlated} = 1 - 1/(1 + exp[-11.5(V_{ref} - 0.57)])$. (d) Probabilistic AND and MUX logics in uncorrelation, implemented with the SNEs and standard AND and MUX logics. The circuits can be reconfigured to implement other probabilistic logics and accommodate varying correlations. (e) Hardware tests of the probabilistic logics in (d) in both uncorrelation and correlation. Bit 0s and 1s are marked in blue and red. For probabilistic MUX, the frequency of select $s$ is half of the inputs to ensure both the inputs participate in the logic operations. The outputs of the probabilistic logics in uncorrelation and correlation are consistent with the statistical relations in Table S1.

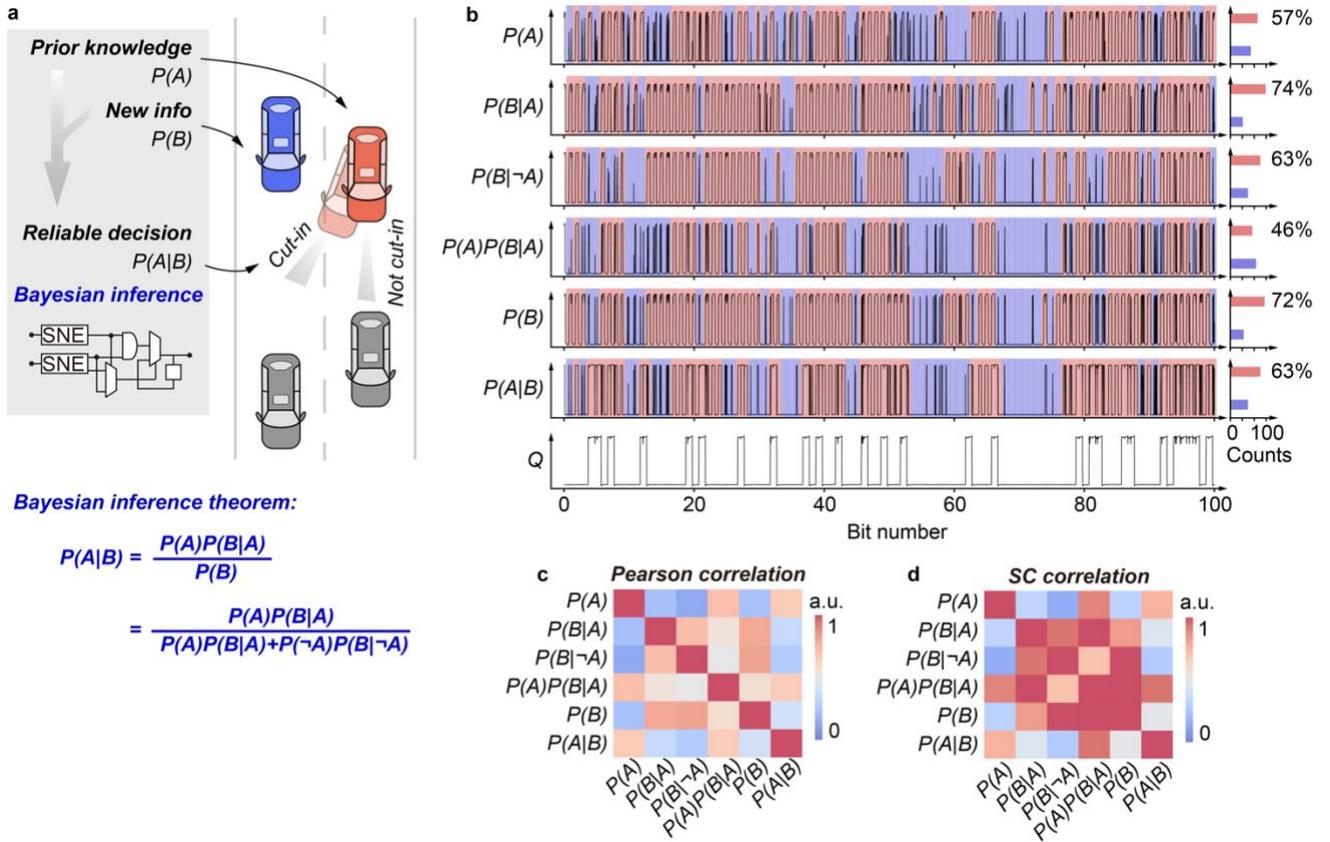

**Figure 3. Hardware implementation of Bayesian inference.** (a) Memristor-enabled Bayesian inference operator and its workflow to perform Bayesian inference for route planning. Prior knowledge with a prior probability $P(A)$ is updated as the new information with probability $P(B)$ becomes available, yielding a reliable decision with a posterior probability $P(A|B)$. Here $P(A)$ represents the initial belief of a vehicle (coloured in red) trying to cut in lanes, and $P(B)$ represents the probability of the vehicle spotting another vehicle (coloured in blue) on the target lane. The Bayesian inference operator is implemented with the memristor-enabled probabilistic AND and MUX logics. The probabilistic logics are used to conduct multiplication and weighted addition operations respectively for functioning as the numerator and denominator conforming to the Bayesian inference theorem. The division is achieved with a probabilistic MUX plus a D-Flip-Flop, following a classic divider design for stochastic numbers, CORDIV (*26*). See Fig. S7 for the detailed circuit design and hardware implementation. (b) Hardware test of the Bayesian inference operator to perform route planning for self-driving vehicles. $P(A)$ and $P(B)$ are manually set to initialise the operator. Stochastic numbers at the key nodes are plotted accordingly. Bit 0s and 1s are marked in blue and red. The result $P(A|B) > P(A)$ means the red vehicle in (a) can cut in lanes with higher confidence. This enhanced decision reliability is attributed to the comprehensive evaluation of both the vehicle itself and the traffic situation. Pairwise (c) Pearson and (d) SC correlations between the stochastic numbers in (b).

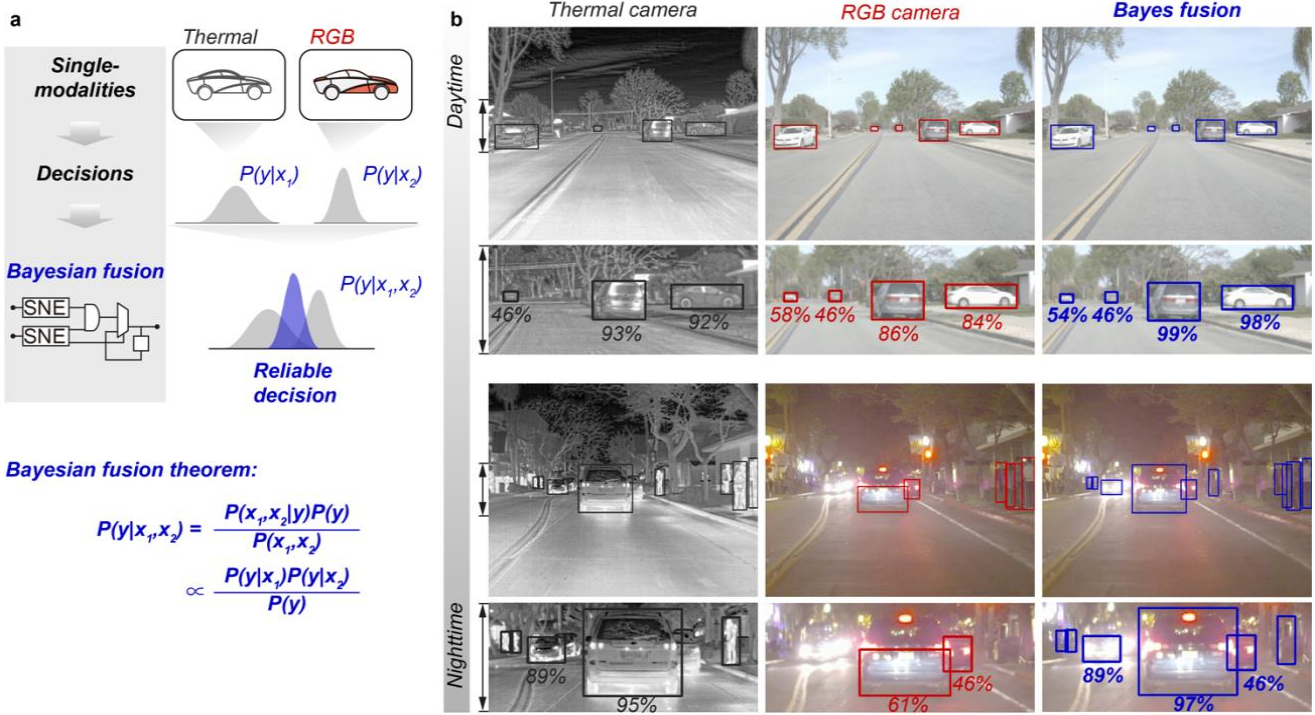

**Figure 4. Hardware implementation of Bayesian fusion.** (a) Memristor-enabled Bayesian fusion operator and its workflow to perform Bayesian fusion. Pre-trained neural networks tailored for single modalities receive RGB and thermal information and output single-modal obstacle detection decisions, denoted as $P(y|x_1)$ and $P(y|x_2)$, where $y$ represents the detected obstacle, and $x_1$ and $x_2$ represent the RGB and thermal information, respectively. The Bayesian fusion operator fuses the single-modal decisions into a reliable decision, denoted as $p(y|x_1, x_2)$, using the Bayesian fusion theorem. The Bayesian fusion operator is implemented with the memristor-enabled probabilistic AND and MUX logics. The probabilistic logics are used to conduct multiplication and weighted addition operations respectively for functioning as the numerator and denominator conforming to the Bayesian fusion theorem. The division is achieved with a probabilistic MUX plus a D-Flip-Flop, following a classic divider design for stochastic numbers, CORDIV (*26*). See Fig. S9 for the detailed circuit design and hardware implementation. (b) Obstacle detection results before and after the Bayesian fusion of the RGB and thermal information under the different visibility conditions during daytime and nighttime. Before fusion, single-modal (RGB or thermal) networks typically lose certain target obstacles and make decisions with insufficient confidence. The Bayesian fusion operator effectively overcomes the single-modal shortages, addresses the target-missing and low-confidence issues, and achieves much more accurate and reliable results.

## Supplementary Figures

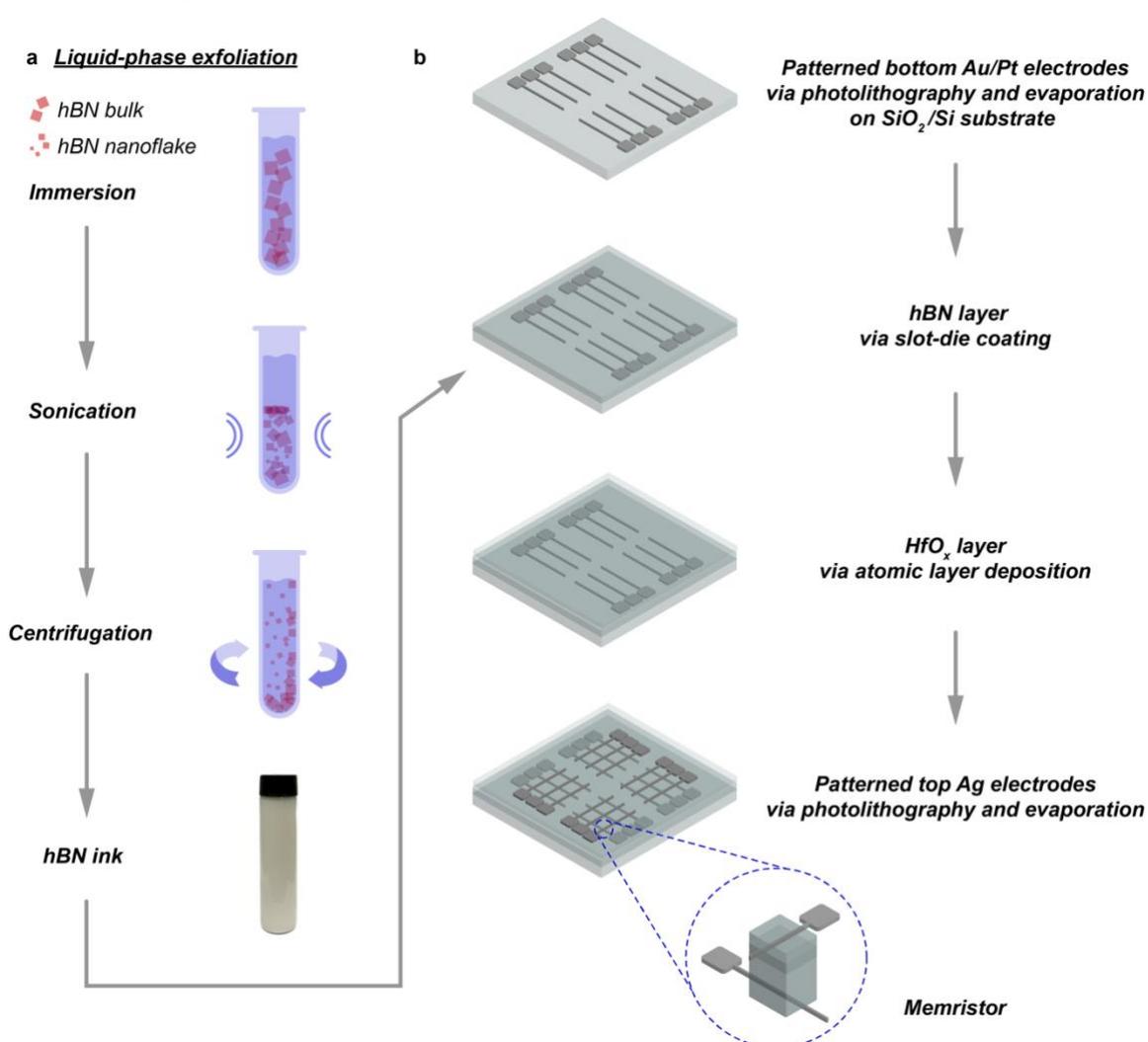

**Figure S1. Memristor fabrication.** (a) Schematic liquid-phase exfoliation of hBN, following our previous report (*1*). After 48-hour bath sonication, the dispersion of the as-exfoliated hBN nanoflakes in isopropanol is centrifuged at 4,000 rpm for 30 minutes (initial concentration of hBN ~10 mg/ml). The supernatant is then carefully decanted, collected, and added with controlled volumes of isopropanol and 2-butanol to formulate an ink in isopropanol/2-butanol (90 vol.%/10 vol.%) with a concentration of ~1 mg/ml. (b) Schematic device fabrication. Starting from a cleansed $SiO_2$/Si substrate, bottom Au/Pt electrodes are deposited and patterned by evaporation and photolithography. hBN layer is deposited on the bottom electrodes by slot-die coating the hBN ink as prepared. $HfO_x$ layer (20 nm) is then deposited using atomic layer deposition to minimise the wash-off of hBN during the subsequent patterning process and increase the fabrication yield to ~100%, while preserving the switching behavior of the filamentary memristors. Next, top silver electrodes are patterned by evaporation and photolithography. As schematically illustrated, the memristors are developed at the cross-points between the top and bottom metal electrodes, in a vertical Au/Pt/hBN/$HfO_x$/Ag device structure.

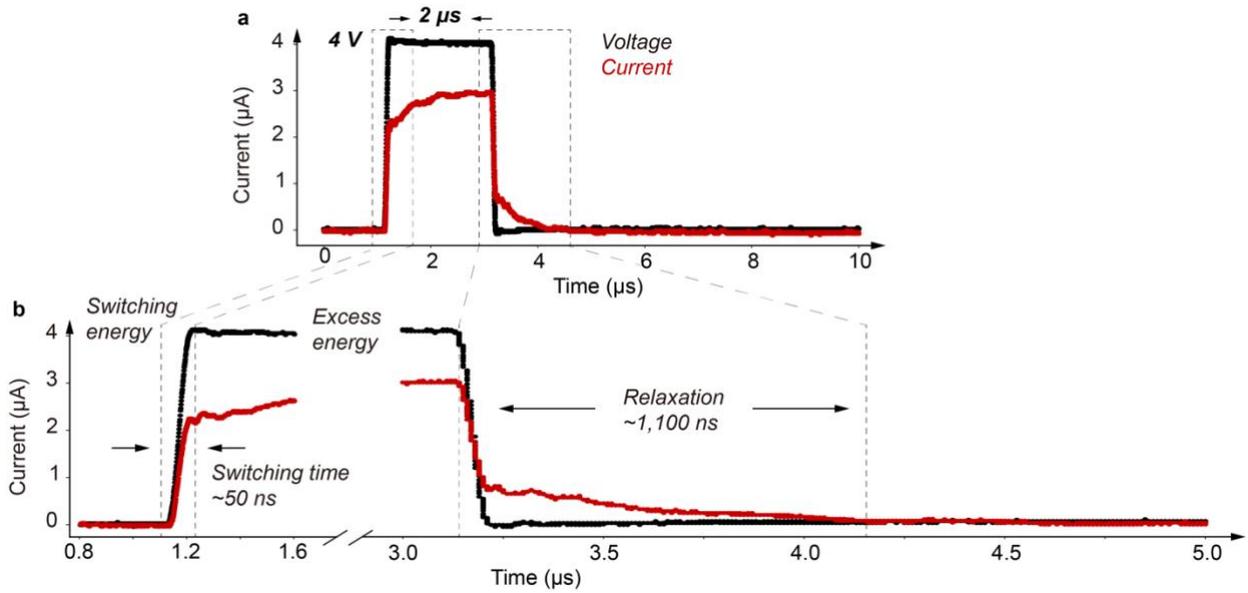

**Figure S2. Switching and energy consumption.** (a) Electrical response of a typical hBN memristor device to 2 μs pulsed signal, showing (b) a switching time of ~50 ns, a switching energy consumption of ~0.16 nJ, and a relaxation time of ~1,100 ns. Given that the entire switching process involves Joule heat, the energy consumption can be segregated into a switching energy and an excess energy (*2*). The switching energy is contributed to the resistive switching process, while the excess energy the energy dissipation following the switching process. The switching energy can be estimated by integrating the product of the transient voltage and current during the switching process, $E = \int_0^t V(t)I(t)dt$.

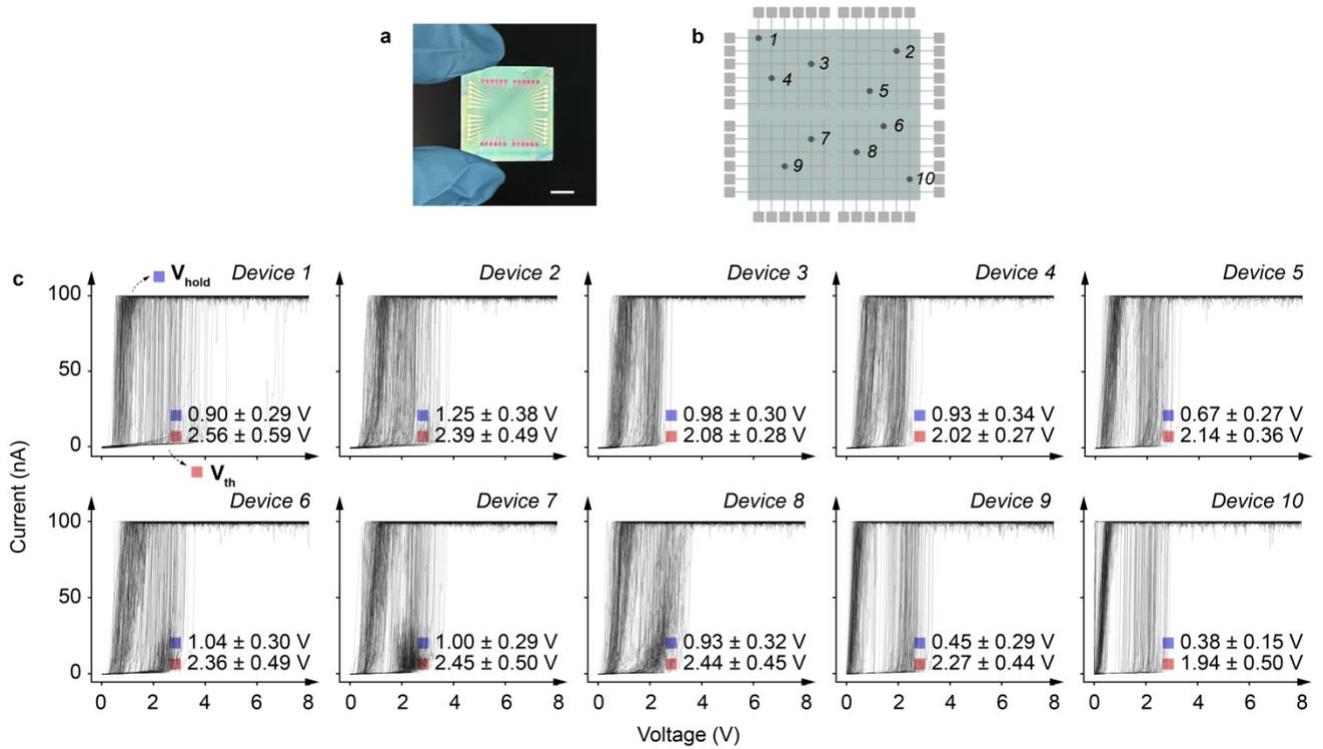

**Figure S3. Sampling test.** (a) 12×12 memristor array in a crossbar configuration, with a fabrication yield approximating 100% (replotted from Fig. 1a), and (b) the corresponding schematic array showing the devices randomly selected for the sampling test. Scale bar – 5 mm. (c) Current-voltage outputs from the randomly selected sampling memristor devices, showing 128-cycle stochastic yet stable switching with a ratio of ~$10^5$ for all the tested devices. A compliance current of 100 nA is set. $V_{hold}$ and $V_{th}$ denote the hold voltage and threshold voltage, respectively.

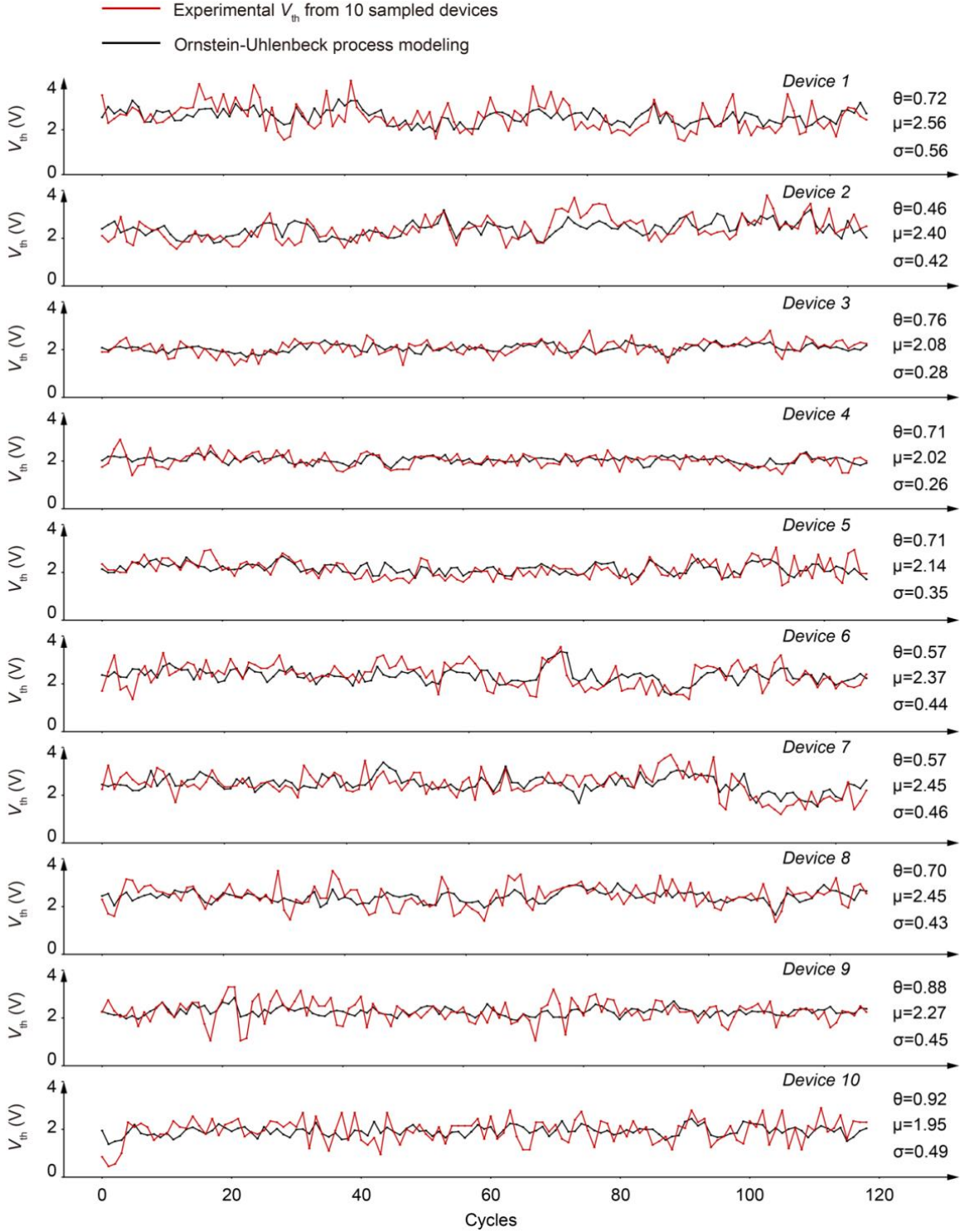

**Figure S4. Stability test of the switching stochasticity from the randomly selected sampling memristor devices.** Ornstein-Uhlenbeck process modeling on the measured threshold voltage $V_{th}$ data points of the 10 sampled memristors across the 128 consecutive sweeping cycles present in Fig. 3c. The experimental $V_{th}$ data points well fit an Ornstein-Uhlenbeck process, $dV_{th,t} = \theta(\mu - V_{th,t}) + \sigma dW_t$, where θ determines the magnitude of mean reversion, μ represents the asymptotic mean, σ stands for the variation, and $dW_t$ denotes the variation of a Wiener process. The Ornstein-Uhlenbeck process describes a stochastic process in a dynamical system (*3*). This proves the stability of the memristor switching stochasticity in prolonged switching operations.

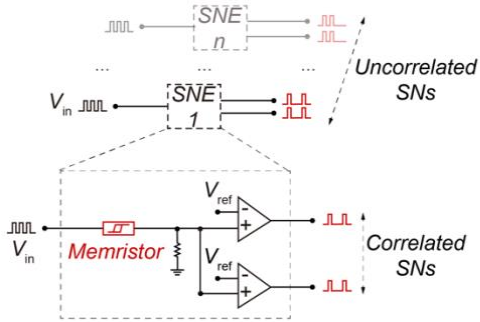 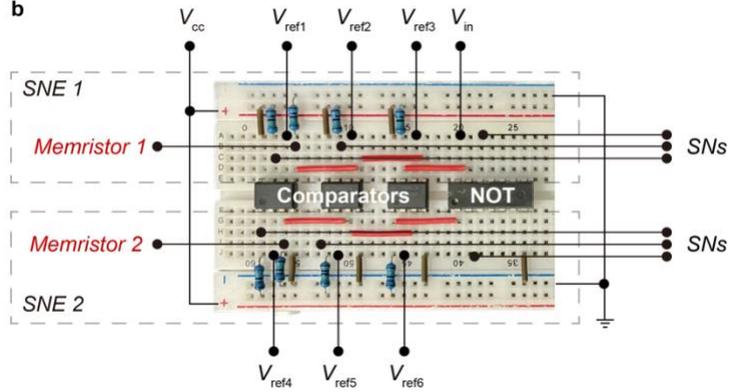

**Figure S5. Stochastic number encoder.** (a) Circuit design (replotted from Fig. 2a) and (b) the experimental setup of stochastic number encoders (SNEs). To build the SNEs, the memristors are tested on a probe station and connected to the logic gates and other electronic components on a breadboard. The electronic components include the comparators and resistors, as well as the NOT gates (connected to the output of comparators if negative correlation is required). Note that the voltage supply of the NOT gates is synchronised with $V_{in}$ to the memristors to avoid output during the pulse intervals.

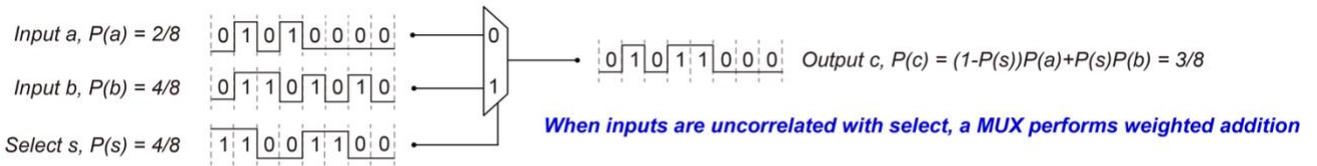
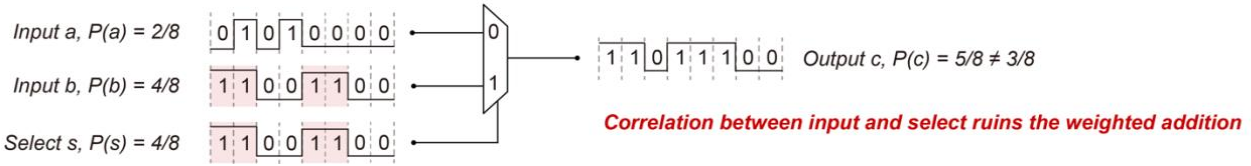

**Figure S6. Probabilistic MUX logic.** (a) An example showing the probabilistic MUX logic as a weighted Adder. In this case, the select $s$ is uncorrelated with the inputs $a$ and $b$. In contrast, the counter example of probabilistic MUX logic in (b) no longer performs weighted addition when the select $s$ is correlated with the inputs $a$ and/or $b$. A positive correlation between input $b$ and select $s$ corrupts the weighted addition operation, as $s$ completely accepts $b$ as part of the output $c$ instead of selecting bits from $b$ with a probability.

**Bayesian inference theorem:**

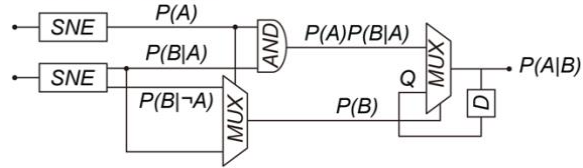

**Operator circuit design:**

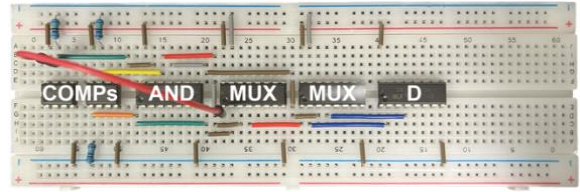

**Hardware implementation:**

![Breadboard photo showing COMPs, AND, MUX, MUX, D components]

**Figure S7. Hardware Bayesian inference operator.** To realise the Bayesian inference theorem (same as that in Fig. 3a), the operator circuit consists of probabilistic AND and MUX logics. They are used to conduct multiplication and weighted addition operations, respectively, for functioning as the numerator and denominator of the Bayesian inference theorem. Probabilistic MUX plus a D-Flip-Flop is used to conduct division, designed following a classic divider design for stochastic numbers, named CORDIV (*4*). During the hardware tests, the memristors (not depicted here for the clarity of the circuit) are tested on a probe station and connected to the logic gates and other electronic components on the breadboard. The electronic components include comparators (COMPs), D-Flip-Flop, and resistors. The comparator reference, voltage supply, and pulsed voltage signal inputs are powered by the arbitrary waveform generators (also not depicted here for clarity).

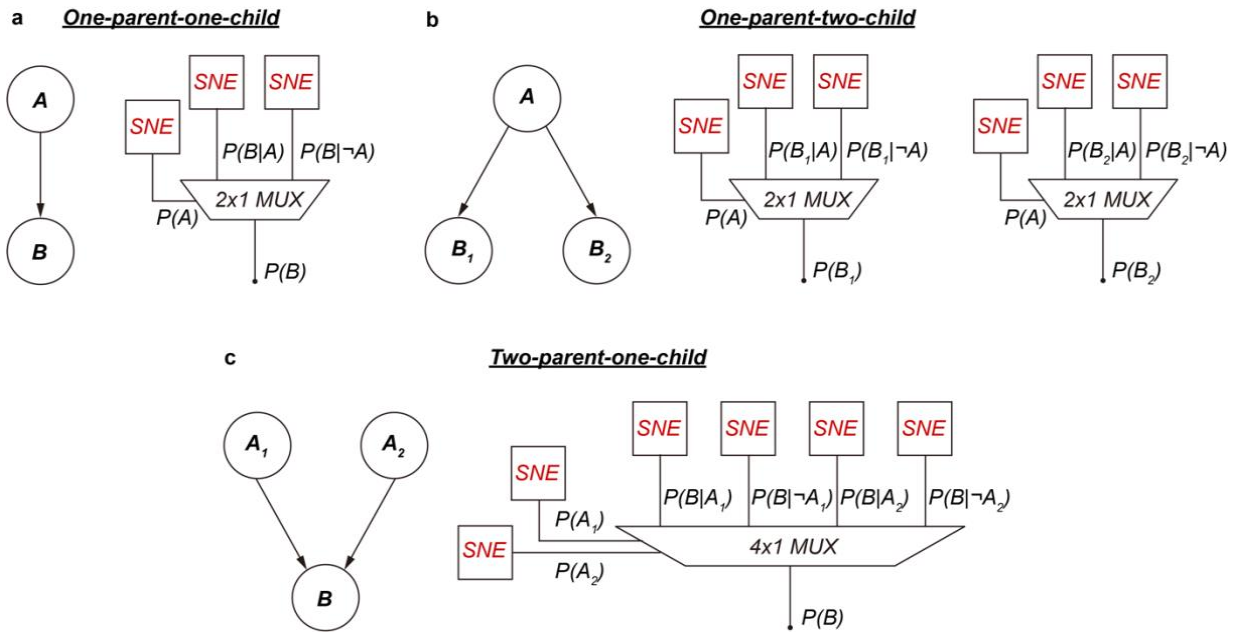

**Figure S8. Basic dependencies within Bayesian inference and corresponding circuit design in probabilistic computing.** Specifically, a 2×1 probabilistic MUX logic can be used in the one-parent-one-child case (a); a 4×1 probabilistic MUX logic can be used in the two-parent-one-child case (b); and two 2×1 probabilistic MUX logics can be used in the one-parent-two-child case (c).

**Bayesian fusion theorem:**

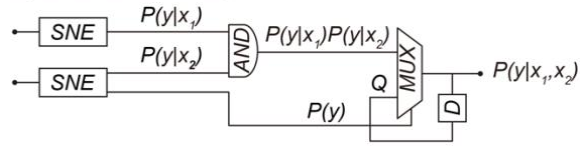

**Operator circuit design:**

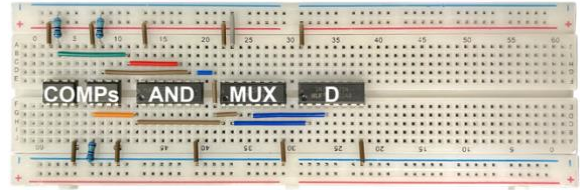

**Hardware implementation:**

**Figure S9. Hardware Bayesian fusion operator.** To realise the Bayesian fusion theorem (same as that in Fig. 4a), the operator circuit consists of probabilistic AND and MUX logics. Probabilistic AND is used to conduct multiplication for functioning as the numerator of the Bayesian fusion theorem. Probabilistic MUX plus a D-Flip-Flop is used to conduct division, designed following a classic divider design for stochastic numbers, named CORDIV (*4*). During the hardware tests, the memristors (not depicted here for the clarity of the circuit) are tested on a probe station and connected to the logic gates and other electronic components on the breadboard. The electronic components include comparators (COMPs), D-Flip-Flop, and resistors. The comparator reference, voltage supply, and pulsed voltage signal inputs are powered by the arbitrary waveform generators (also not depicted here for clarity).

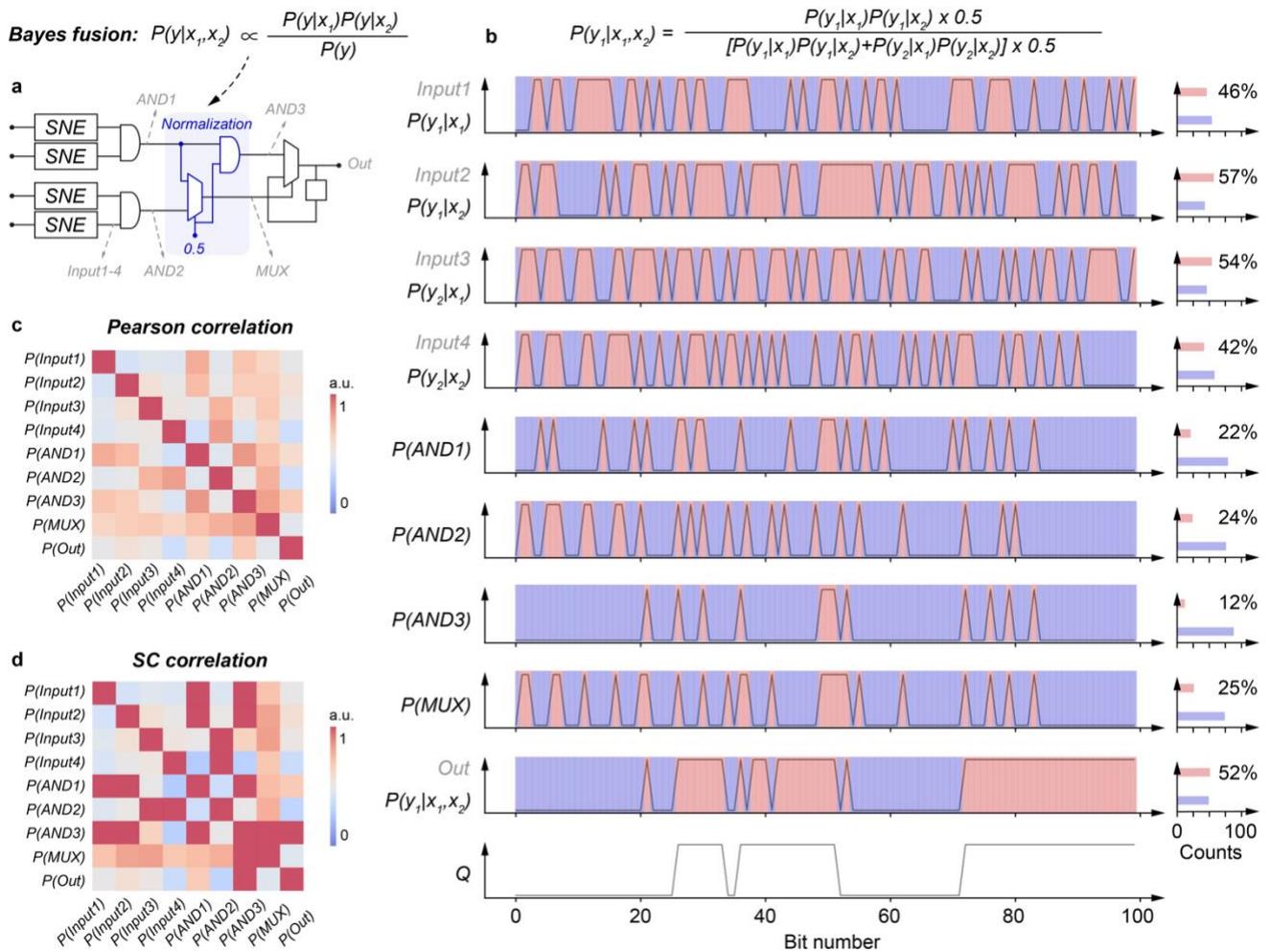

**Figure S10. Bayesian fusion operator with a normalization module.** (a) Circuit design of a Bayesian fusion operator. Probabilistic AND and MUX logics are used to conduct multiplication and weighted addition operations respectively for functioning as the numerator and denominator for the Bayesian fusion theorem. Probabilistic MUX plus a D-Flip-Flop is used to conduct division, designed following a classic divider design for stochastic numbers, named CORDIV (*4*). Additionally, a normalization module is integrated. (b) Simulation result of Bayesian fusion using the normalization module. The stochastic numbers at the key nodes in (a) are plotted. Bit 0s and 1s are marked in blue and red. Pairwise (c) Pearson and (d) SC correlations between stochastic numbers in (b).

**Supplementary Table**

**Table S1. Probabilistic logic operations in varying correlation conditions.** Probabilistic AND, OR, XOR, and MUX logics in the uncorrelated, positive correlated, and negative correlated cases are presented. The stochastic numbers are assumed in a unipolar format (5).

|  | Uncorrelated | Positively correlated | Negatively correlated |
|---|---|---|---|
| AND | $P(c) = P(a)P(b)$ | $P(c) = min(P(a), P(b))$ | $P(c) = max(P(a) + P(b) - 1, 0)$ |
| OR | $P(c) = P(a) + P(b) - P(a)P(b)$ | $P(c) = max(P(a), P(b))$ | $P(c) = min(1, P(a) + P(b))$ |
| XOR | $P(c) = P(a) + P(b) - 2P(a)P(b)$ | $P(c) = \|P(a) - P(b)\|$ | $P(c) = P(a) + P(b)$, if $P(a) + P(b) \leq 1$; $P(c) = 2 - (P(a) + P(b))$, otherwise. |
| MUX | \multicolumn{3}{c}{$P(c) = (1 - P(s))P(a) + P(s)P(b)$, if $s$ is uncorrelated with $a$ and $b$} | | |

**Supplementary References**

1. L. Song, P. Liu, J. Pei, F. Bai, Y. Liu, S. Liu, Y. Wen, L. W. T. Ng, K. P. Pun, S. Gao, M. Q. H. Meng, T. Hasan, G. Hu, Spiking neurons with neural dynamics implemented using stochastic memristors. *Advanced Electronic Materials* **2300564**, 1–9 (2023).
2. S. S. Teja Nibhanupudi, A. Roy, D. Veksler, M. Coupin, K. C. Matthews, M. Disiena, Ansh, J. V. Singh, I. R. Gearba-Dolocan, J. Warner, J. P. Kulkarni, G. Bersuker, S. K. Banerjee, Ultra-fast switching memristors based on two-dimensional materials. *Nature Communications* **15** (2024).
3. S. Dutta, G. Detorakis, A. Khanna, B. Grisafe, E. Neftci, S. Datta, Neural sampling machine with stochastic synapse allows brain-like learning and inference. *Nature Communications* **13**, 2571 (2022).
4. T. H. Chen, J. P. Hayes, Design of division circuits for stochastic computing. *Proceedings of IEEE Computer Society Annual Symposium on VLSI, ISVLSI* **2016-Septe**, 116–121 (2016).
5. A. Alaghi, W. Qian, J. P. Hayes, The promise and challenge of stochastic computing. *IEEE Transactions on Computer-Aided Design of Integrated Circuits and Systems* **37**, 1515–1531 (2018).